\documentclass[10pt]{article} 
\usepackage[preprint]{tmlr}
\usepackage{graphicx}
\usepackage{booktabs}
\usepackage{listings}
\usepackage{algorithm}
\usepackage{algpseudocode} 

\usepackage{amsmath,amsfonts,bm}

\def\eqref#1{equation~\ref{#1}}

\def\1{\bm{1}}

\DeclareMathAlphabet{\mathsfit}{\encodingdefault}{\sfdefault}{m}{sl}
\SetMathAlphabet{\mathsfit}{bold}{\encodingdefault}{\sfdefault}{bx}{n}

\usepackage{hyperref}
\usepackage{url}

\title{R-DEIM Net: An Efficient Rationale-Augmented Dual-Expert Interaction Model for Paraphrase Detection}

\author{\name Pushp \email pushp.g23@iiits.in \\
      \addr Indian Institute of Information Technology (IIIT), Sri City, India
      \AND
      \name Vaibhav Prajapati \email vaibhav.prajapati@utn.de \\
      \addr University of Technology Nuremberg (UTN), Germany
      \AND
      \name Himangshu Sarma \email himangshu.sarma@iiits.in \\
      \addr Indian Institute of Information Technology (IIIT), Sri City, India}

\def\openreview{\url{https://openreview.net/forum?id=XXXX}} 

\begin{document}

\maketitle

\begin{abstract}
Recent advances in paraphrase detection reveal a fundamental trade-off: large language models achieve high accuracy but require high computation, while efficient Siamese-BERT variants offer practical scalability with reduced transparency in rationale generation. We present R-DEIM Net, a 76M-parameter dual-expert architecture exploring whether moderate-scale models can achieve competitive accuracy on paraphrase detection while enabling human-readable rationale generation. The architecture combines two specialized components: an Interaction Expert that captures token-level similarity patterns through multi-scale 2D convolutions and attention head allowing variable input length, and a Reasoning Expert that uses a Flan-T5-small decoder to generate rationales as auxiliary supervision. Rather than re-encoding generated text, we extract and pool decoder hidden states as complementary features for classification. On the Quora Question Pairs dataset, R-DEIM Net achieves 90.07\% accuracy and 90.16\% F1-score via 10-fold cross-validation. This represents competitive performance with strong transformer-based baselines (e.g., MFAE BERT: 90.54\% accuracy) and recent large language model based approaches (LLaMA-70B) while using a substantially smaller parameter budget. The model generates rationales alongside predictions, providing potential for auxiliary human-readable descriptions.
\end{abstract}

\section{Introduction}
\label{sec:introduction}

Accurately discerning semantic equivalence between text segments for paraphrase detection is a cornerstone of modern Natural Language Understanding (NLU) \citep{importance_of_paraphrase_detection_1,importance_of_paraphrase_detection_2}. This capability is critical for applications ranging from intelligent search and conversational AI to content de-duplication \citep{de_duplication} and plagiarism detection \citep{machine_translation_1,machine_translation_2,natural_language_generation,Question_answering_1,Question_answering_2,plagirism}. Following the 2017 Kaggle competition \citep{kaggle_comp}, the Quora Question Pairs (QQP) dataset has become the standard benchmark for this task, requiring models to capture semantic identity across diverse surface-level wordings \citep{beyond_lexical_1,beyond_lexical_2}.

Recent progress in paraphrase detection reveals distinct trade-offs between competing objectives. High performance is achieved by Large Language Models (LLMs) \citep{70b_sota} and ensemble BERT \citep{mfae}, but these models result in high computational costs limiting real-world deployment. Conversely, efficient Siamese-BERT architectures \citep{siaseme_approach_1,stanford} achieve practical scalability but provide only binary predictions with limited transparent rationale generation. While attention mechanisms can partially address this gap in mathematical aspect, most efficient paraphrase detection systems still lack explicit, rationales explaining their decisions in human-readable format.

To explore whether moderate-scale models can achieve competitive accuracy while providing synthetic rationale generation, this paper introduces R-DEIM Net (Rationale-Augmented Dual-Expert Interaction Model). R-DEIM Net takes a sentence pair as input and outputs a binary classification alongside generated text rationales. The architecture combines two complementary experts: (1) an \textit{Interaction Expert} that captures token-level similarity patterns through multi-scale 2D convolutions and attention-based pooling to handle variable-length inputs, and (2) a \textit{Reasoning Expert} that uses Flan-T5-small decoder to generate rationales as auxiliary training supervision. The Reasoning Expert's hidden states during rationale generation act as a feature vector complementing the interaction patterns for final classification.

The primary contributions are as follows:
\begin{itemize}
  \item We propose R-DEIM Net, a dual-expert framework that jointly models token-to-token interactions and generative reasoning signals for paraphrase detection.
  
  \item We design an interaction expert that uses 2D convolutions and an attention-based pooling strategy to efficiently capture token alignment patterns from variable length sentence pairs.

  \item By using decoder hidden states from a generative model as latent reasoning features, we adopt a mechanism that grounds classification decisions in natural language features without re-encoding generated rationales, which avoids unnecessary computational overhead and additional encoding layers.

  \item We present a thorough empirical analysis to show how interaction and reasoning components contribute to semantic decision-making, including structural alignment, kernel activation behavior, and POS-based perturbation.

\end{itemize}

\section{Related Work}

Paraphrase detection on the Quora Question Pairs (QQP) dataset has evolved from traditional lexical features to deep sequential architectures and, recently, to massive LLMs. This progression reflects a shift from manual feature engineering toward automated semantic representation, which requires trade-offs in efficiency and transparent rationale generation.
\paragraph{Traditional and Lexical Approaches:}
Early methodologies relied on manual feature engineering and basic vectorization, with accuracies typically ranging from 63\% to 85\%. \citet{literature_review_1} demonstrated that a simple Continuous Bag of Words (CBOW) model achieved 83.4\% accuracy, while \citet{literature_review_2} reached 82.44\% using 28 hand-crafted features with an XGBoost classifier. Similar lexical strategies were explored in ABCNN \citep{abcnn}, PWIM \citep{pwim}, and the ``cascadedCN'' model \citep{literature_review_3}. Techniques like Lexical Decomposition and Composition (LDC) \citep{ldc} and Syn-tree \citep{esim} attempted to capture alignment by breaking sentences into primitive components. While computationally light, these methods are limited by the inability to capture deep semantic context or complex word-order dependencies.

\paragraph{Sequential and Convolutional Architectures:}
The shift toward neural sequence modeling introduced RNN and CNN-based frameworks, yielding accuracies between 79\% and 89\%. Convolutional approaches, such as Siamese-CNN and MP-CNN \citep{bimpm}, utilized spatial filters to identify local n-gram similarities. Concurrently, recurrent models like Siamese-LSTM, MP-LSTM \citep{bimpm}, and ESIM \citep{esim} focused on long-range dependencies. Universal encoders like InferSent \citep{infersent} and GenSen \citep{gensen} sought general representations, while SSE \citep{sse}, CAS-LSTM, and Bi-CAS-LSTM \citep{cas} introduced stacked architectures to refine context. Performance boundaries were further pushed using Bi-LSTM and attention framework by \citet{smote_data}. This model leveraged SMOTE \citep{smote} for augmentation, despite prevailing concerns that oversampling in the feature space may fail to preserve semantic validity in textual representations \citep{smote_1, smote_2}. Concurrent systems like pt-DecAtt \citep{att} and LSTM+ElBiS \citep{elbis} demonstrated similar gains in predictive power. However, these models often struggle with variable-length inputs and lack transparency in decision-making logic.

\paragraph{High-Interaction and Transformer-based Systems:}
To capture complex semantic interactions, models operating over dense interaction spaces were developed, with accuracies now reaching the 88\% to 90\% range. DIIN \citep{diin} and Multiway Attention Networks (MwAN) \citep{mwan} utilize sophisticated attention to align sentence pairs. The integration of ensemble BERT led to MFAE \citep{mfae, regmapr} and CDTFME-Aver \citep{aver}, which leverage BERT and ELMo embeddings. At the extreme scale, the most recent model, LLaMA-70B, is applied by \citet{70b_sota}.

\begin{figure}[t]
    \centering
    \includegraphics[width=0.9\textwidth]{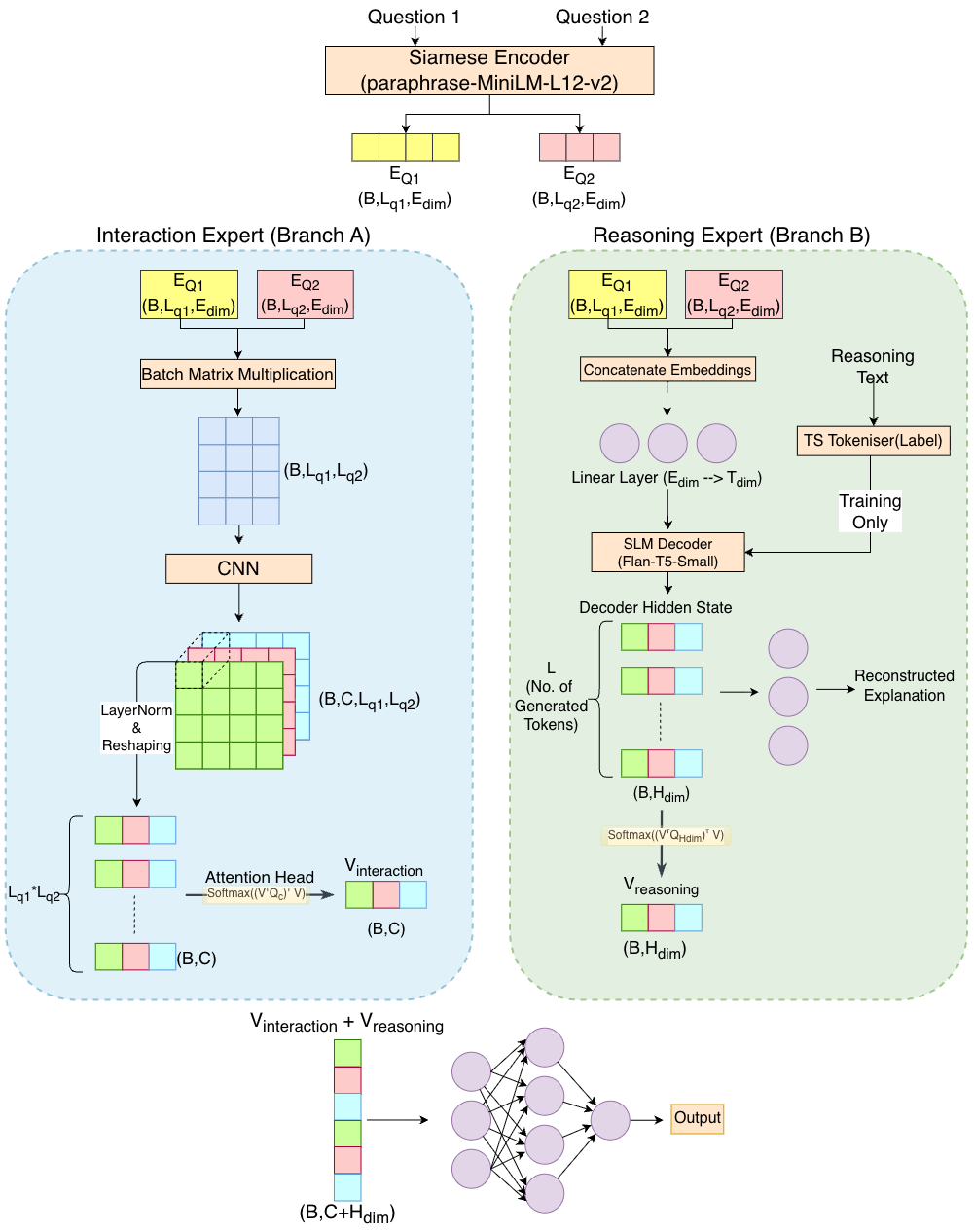}
    \caption{\textbf{R-DEIM Net Architecture.} A dual-expert multi-task framework. Branch A (Interaction) captures token-to-token patterns via multi-scale 2D-CNNs, while Branch B (Reasoning) distills latent decoder trajectories into a ``reasoning vector'' to ground classification in natural language rationales.}
    \label{fig:main_architecture}
\end{figure}

\paragraph{Technical Gaps:}
A critical analysis of existing literature reveals three primary deficiencies. First, a significant portion of prior work focuses almost exclusively on raw accuracy, frequently failing to report $F_1$-score, which leaves performance on imbalanced data unverified. Second, nearly all existing architectures rely on a fixed \textit{max\_tokens} limit. This forces a compromise between truncating semantically rich sentences or introducing excessive padding, both of which create sparse, inefficient matrices and noisy attention weights. Finally, despite their predictive power, these models provide binary labels.



\begin{table}[t]
\centering
\caption{Encoder benchmarking on QQP.}
\label{tab:encoder_comparison}
\begin{tabular}{lccccc}
\toprule
\textbf{Model} & \textbf{Size} & \textbf{Train Acc.} & \textbf{Train $F_1$} & \textbf{Test Acc.} & \textbf{Test $F_1$} \\
\midrule
bert-base-uncased \\ \citep{bert_base_uncased}
& 110M & $90.95 \pm 10.04$ & $89.54 \pm 14.49$ & $85.99 \pm 8.11$ & $84.63 \pm 12.64$ \\
paraphrase-albert-base-v2
& 12M & $88.29 \pm 6.26$ & $87.87 \pm 6.74$ & $85.00 \pm 3.99$ & $84.49 \pm 4.43$ \\
paraphrase-MiniLM-L3-v2
& 17M & $90.55 \pm 1.53$ & $90.61 \pm 1.53$ & $86.52 \pm 0.57$ & $86.62 \pm 0.58$ \\
paraphrase-MiniLM-L6-v2
& 22M & $91.16 \pm 1.69$ & $91.25 \pm 1.67$ & $87.05 \pm 0.71$ & $87.20 \pm 0.66$ \\
\textbf{paraphrase-MiniLM-L12-v2}
& \textbf{33M} & $\mathbf{92.97 \pm 1.30}$ & $\mathbf{93.05 \pm 1.27}$ & $\mathbf{88.42 \pm 0.37}$ & $\mathbf{88.56 \pm 0.35}$ \\
paraphrase-mpnet-base-v2
& 110M & $88.18 \pm 13.35$ & $85.36 \pm 19.36$ & $84.08 \pm 11.09$ & $81.31 \pm 17.15$ \\
paraphrase-TinyBERT-L6-v2
& 67M & $92.95 \pm 1.67$ & $93.01 \pm 1.65$ & $88.59 \pm 0.64$ & $88.70 \pm 0.61$ \\
\bottomrule
\end{tabular}
\end{table}

\section{Method}
\label{sec:methodology}
Question matching can be viewed as a classification task that seeks a label $y \in \mathcal \{\text{Duplicate, Non-Duplicate}\}$ for a given question pair. Figure~\ref{fig:main_architecture} illustrates our approach for this task to classify and provide rationale. In the following, we describe the individual component of this approach.

\subsection{Branch A: Interaction Expert}
This branch is designed to move beyond simple vector similarity and analyze token-to-token semantic relationships between the two questions. Both input questions are first processed by the shared encoder. To select the optimal encoder, several pre-trained encoders were evaluated (Table \ref{tab:encoder_comparison}), and \texttt{paraphrase-MiniLM-L12-v2} \citep{sentence_transformer, sentence_transformer_are_good} was selected. While models like TinyBERT achieved marginally higher raw metrics, MiniLM-L12 offers a superior trade-off, delivering competitive accuracy (88.42\%) and F1-scores (88.56\%) with a significantly lesser parameter ($\approx$ 33M).

Next, a 2D interaction matrix $M$ is computed to capture the token-by-token similarity \citep{dot_product_1,dot_product_2} between every token in $Q_1$ and $Q_2$:
\begin{equation}
    M = E_{Q1} \cdot E_{Q2}^T
    \quad \text{where } M \in \mathbb{R}^{B \times L_{q1} \times L_{q2}}
\end{equation}

Several parallel 2D convolution layers with varying kernel sizes  are applied on $M$ to discover local interaction patterns. A $n \times n$ kernel can identify patterns of similarity across n-token phrases (semantic n-grams) \citep{Gram_cnn}, capturing features that a simple 1-to-1 dot product would miss. The resultant feature map has shape ($B, C, L_{q1}, L_{q2}$).

To create a fixed-size vector out of variable input, the feature map is flattened into a sequence $X$ and processed with a custom \texttt{Attention Head}. A trainable query vector $v_q \in \mathbb{R}^{C}$ scores the importance of each feature $X_i$ by computing alignment scores $s$:
\begin{equation}
    s_i = (X_i \cdot v_q) + b_{imp}
\end{equation}
Here, $b_{imp}$ is a small, scalar bias (the \texttt{importance\_factor}) providing a baseline relevance score. These scores are normalized into attention weights using a Softmax function.
Finally, the fixed-size output vector $v_{interaction}$ is computed as the weighted sum of all feature vectors:
\begin{equation}
    v_{interaction} = \left( \sum_{i=1}^{L_{flat}} \frac{\exp(s_i)}{\sum_{j=1}^{L_{flat}} \exp(s_j)} \right) \cdot X_i
\end{equation}
This attention-pooling mechanism effectively distills variable-length interaction patterns into a single, fixed-size vector for the classifier.

\subsection{Branch B: Differentiable Reasoning Expert}
This branch serves a dual purpose: it generates an explicit rationale for the model's decision and converts the latent logic itself into a feature vector. \texttt{Flan-T5-small} decoder \citep{t5} is employed, where the token embeddings $E_{Q1}$ and $E_{Q2}$ from the shared encoder are provided as `encoder hidden states`, giving the decoder full context. To bridge the dimension gap between the siamese encoder (384) and the T5 decoder (512), a simple linear projection layer is used.

During training, while generating a rationale the decoder is tasked with auto-regressive decoding with teacher forcing \citep{attention_is_all_you_need}. Rather than re-encoding the generated surface text which would require passing the text through another encoder, adding unnecessary computational overhead and training complexity the sequence of decoder hidden states, $H_{dim}$, is directly extracted as each token is generated.This results in a variable-length tensor ($L \times H_{dim}$) representing the reasoning. Similar to Branch A, a separate `Attention Head' is applied to pool this sequence into a single, fixed-size reasoning vector, $v_{reasoning}$.

\subsection{Fusion and Multi-Task Training Objective}
The final classification combines insights from both experts. The two feature vectors are concatenated to form a final representation ($v_{final}$) containing both interaction patterns and generative logic which is passed through an MLP.

The model is optimized on a weighted multi-task loss combining classification ($L_{classify}$) and generation ($L_{generate}$). $L_{generate}$ is the standard \texttt{CrossEntropyLoss} from the T5 decoder. For classification, false positives are costly, as in duplicate detection incorrectly merging distinct questions is more detrimental to user experience than missing a duplicate. A per-sample weight $W_i$ is defined based on the false positive penalty $w_{fp}$:
\begin{equation}
    W_i = \begin{cases} w_{fp} & \text{if } y_i = 0 \\ 1 & \text{if } y_i = 1 \end{cases}
\end{equation}
The classification loss is the weighted mean of the per-sample BCE losses:
\begin{equation}
    L_{classify} = \frac{1}{N} \sum_{i=1}^{N} W_i \cdot \Big( -[y_i \log(\sigma(\hat{y}_{logits\_i})) + (1 - y_i) \log(1 - \sigma(\hat{y}_{logits\_i}))] \Big)
\end{equation}
The final multi-task loss trained by the optimizer is:
\begin{equation}
    L_{total} = \alpha \cdot L_{classify} + (1 - \alpha) \cdot L_{generate}
\end{equation}

\section{Experiments}
This section presents the empirical evaluation of R-DEIM Net, including the experimental setup, comparative performance on the Quora Question Pairs (QQP) benchmark, and a multi-level analysis of model behavior. All experiments were conducted on a system equipped with an NVIDIA Quadro RTX 6000 GPU, 64 GB RAM, and an Intel Xeon CPU.

\subsection{Experimental Setup}
\paragraph{Dataset Curation and Preprocessing}
\label{sec:dataset}
Evaluation was performed on the QQP dataset \citep{kaggle_comp}, consisting of 404,290 pairs. To train the Reasoning Expert, the dataset was augmented with synthetic rationales generated by an LLM using a few-shot Chain of Thought (CoT) prompt \citep{cot} to ensure high-quality reasoning (Appendix \ref{sec:appendix_rationale} and \ref{sec:semantic_consistency}). Following preprocessing, the final dataset comprised 400,520 annotated pairs.

\begin{algorithm}[!t]
\caption{R-DEIM Net Multi-Task Training with Decoupled Weight Decay}
\label{algo}
\begin{algorithmic}[1]
\State \textbf{Input:} Training set $\{(Q_{1,i}, Q_{2,i}, y_i, R_i)\}_{i=1}^N$ 
\State \textbf{Parameters:} Encoder $\theta_{enc}$, Experts $\theta_{A}, \theta_{B}$, Classifier $\theta_{c}$ 
\State \textbf{Hyper-parameters:} Multi-task weight $\alpha$, Learning rates $\eta$, weight decay $\lambda$
\Repeat
    \State Sample mini-batch of size $m$ 
    \State $E_{Q1}, E_{Q2} \gets \text{MiniLM-L12}(Q_1, Q_2; \theta_{enc})$ 
    \State $v_i \gets \text{Attn}(\text{CNN}(E_{Q1} \cdot E_{Q2}^\top; \theta_A))$ 
    \State $H_{dec}, \mathcal{L}_{gen} \gets \text{T5}(R | E_{Q1}, E_{Q2}; \theta_B)$ 
    \State $v_r \gets \text{Attn}(H_{dec})$ 
    \State $\hat{y} \gets \text{MLP}([v_i; v_r]; \theta_c)$ 
    \State $\mathcal{L}_{total} = \alpha \mathcal{L}_{BCE}(\hat{y}, y) + (1-\alpha) \mathcal{L}_{gen}$ 
    \State $g_t \gets \nabla_{\theta} \mathcal{L}_{total}$ 
    \State $m_t, v_t \gets \text{Update moments}(g_t)$ 
    \State $\theta_{t+1} \gets \theta_t - \eta \left( \frac{m_t}{\sqrt{v_t} + \epsilon} + \lambda \theta_t \right)$ 
\Until{$\theta$ converges} 
\end{algorithmic}
\end{algorithm}

\paragraph{Training Details and Parameters}
\label{sec:implementation_details}
R-DEIM Net employs a paraphrase-MiniLM-L12-v2 encoder (384-dim) and a Flan-T5-small decoder (512-dim), connected via a linear projection layer. The base Flan-T5-small model contains 77M parameters, split between a 35.3M parameter encoder and a 41.6M parameter decoder. Although these components are typically integrated, our architecture utilizes only the decoder for rationale generation, leaving the T5 encoder unused and excluded from the training process. Lightweight task-specific heads introduce fewer than 1M additional parameters. Specifically, the total parameter count includes the T5 decoder (41.6M), the MiniLM encoder (33M), and custom projection/attention heads ($\approx$ 1M), totaling roughly 76M parameters.

Training was performed (\textbf{Algorithm \ref{algo}}) for a fixed number of epochs using AdamW optimizer \citep{adamw} with different learning rates \cite{learning_rate} for pre-trained and task-specific components. All architectural choices are summarized in Appendix~\ref{sec:Hyperparamters}. The classification objective uses a weighted binary cross-entropy loss that explicitly penalizes false positives by assigning a higher weight to the majority \textit{ non-duplicate} class. This design encourages conservative duplicate predictions, ensuring that a sentence pair is labeled as \textit{Duplicate} only when strong semantic evidence is present. Reproducibility is ensured through the use of standardized, publicly available pre-trained models and the explicit reporting of all training hyperparameters in Appendix~\ref{sec:Hyperparamters}.

\subsection{Experimental Results}
\label{sec:results}

Table \ref{model_comparison_table} presents the 10-fold cross-validation results. R-DEIM Net achieved an average $F_1$-Score of 90.16\% and accuracy of 90.07\%. This performance improves on the efficient baselines listed (e.g., Bi-LSTM \citep{smote_data}) and is within reported ranges for large LLMs in prior work \citep{70b_sota} while being $\approx$ 900x smaller. To ensure the integrity of the evaluation, primary sources of data leakage were investigated by performing the 10-fold cross-validation split prior to any preprocessing. This ensured that lexical statistics and rationale patterns from validation sets were never visible during training. 

Crucially, the full R-DEIM Net outperformed the strong ``Interaction-Only'' baseline (Branch A only), which achieved 88.63\% $F_1$ and 88.51\% accuracy. The addition of the Reasoning Expert (Branch B) provided a substantial +1.53\% boost in $F_1$-Score. This confirmed that the generative ``reasoning vector'' captures semantic nuances that the interaction matrix alone misses, effectively replacing the need for manual feature engineering.

\begin{table}[t]
\centering
\caption{\textbf{Comparative Performance.} R-DEIM Net obtains strong accuracy and $F_1$ on QQP using a compact architecture (76M params).}
\label{model_comparison_table}

\begin{tabular}{lcc}
\toprule
\textbf{Model} & \textbf{Acc.} & \textbf{$F_1$-Score} \\
\midrule
ABCNN \citep{abcnn} & 63.59\% & - \\
Syn-tree \citep{esim} & 75.50\% & - \\
Siamese-CNN \citep{bimpm} & 79.60\% & - \\
MP-CNN \citep{bimpm} & 81.38\% & - \\
XGBoost \citep{literature_review_2} & 82.44\% & 80.44\% \\
Siamese-LSTM \citep{bimpm} & 82.58\% & - \\
MP-LSTM \citep{bimpm} & 83.21\% & - \\
Bi-LSTM \citep{smote_data} & 83.30\% & 87.00\% \\
CBOW \citep{literature_review_1} & 83.40\% & 77.80\% \\
PWIM \citep{pwim} & 83.40\% & - \\
$ESIM_{Syn+tree}$ \citep{esim} & 85.40\% & - \\
LDC \citep{ldc,bimpm} & 85.55\% & - \\
ESIM \citep{esim} & 85.00\% & - \\
InferSent \citep{infersent} & 86.60\% & - \\
GenSen \citep{gensen} & 87.01\% & - \\
LSTM+ElBiS \citep{elbis} & 87.30\% & - \\
pt-$DecAtt_{word}$ \citep{att} & 87.54\% & - \\
SSE \citep{sse} & 87.80\% & - \\
CDTFME-Aver \citep{aver} & 88.00\% & - \\
pt-$DecAtt_{char}$ \citep{att} & 88.40\% & - \\
CAS-LSTM \citep{cas} & 88.40\% & - \\
Bi-CAS-LSTM \citep{cas} & 88.60\% & - \\
REGMAPR \citep{regmapr} & 88.64\% & - \\
BiMPM \citep{bimpm} & 88.17\% & - \\
DIIN \citep{diin} & 89.06\% & - \\
MwAN \citep{mwan} & 89.12\% & - \\
LLaMA-7B \citep{70b_sota} & 89.10\% & 71.90\% \\
MFAE (ELMo) \citep{mfae} & 89.61\% & - \\
MFAE (BERT) \citep{mfae} & 89.79\% & - \\
DIIN (Ensemble) \citep{diin} & 89.84\% & - \\
LLaMA-70B \citep{70b_sota} & 90.30\% & 74.40\% \\
MFAE (BERT Ens.) \citep{mfae} & 90.54\% & - \\
\midrule
R-DEIM Net (Interaction-only) & 88.42\% & 88.56\% \\
\textbf{R-DEIM Net (Proposed)} & \textbf{90.07\%} & \textbf{90.16\%} \\
\bottomrule
\end{tabular}
\end{table}

\begin{table}[t]
\centering
\caption{Transfer learning results of R-DEIM Net.}
\label{tab:transfer_learning}

\begin{tabular}{lccccc}
\toprule
&
\multicolumn{2}{c}{Zero-shot}
&
\multicolumn{2}{c}{Fine-tuning} \\
\cmidrule(lr){2-3}
\cmidrule(lr){5-6}

\textbf{Dataset}
& \textbf{Acc.}
& \textbf{$F_1$}
& \textbf{Epochs}
& \textbf{Acc.}
& \textbf{$F_1$} \\
\midrule

SoDD \citep{sodd}
& 73.50 & 63.01 & 3 & 88.60 & 88.40 \\

PIT \citep{pit}
& 68.97 & 70.17 & 3 & 81.26 & 81.04 \\

PAWS \citep{paws}
& 53.45 & 53.54 & 2 & 68.80 & 67.97 \\

MRPC \citep{mrpc}
& 50.92 & 50.29 & 3 & 69.51 & 65.85 \\

SciTail \citep{scitail}
& 66.60 & 60.60 & 2 & 81.47 & 81.59 \\

SprintFAQ \citep{sprintfaq_1} \\
\citep{sprintfaq_2, sprintfaq_3}
& 97.90 & 98.15 & 2 & 99.38 & 99.41 \\

CQADupStack \citep{cqadup}
& 52.34 & 39.24 & 4 & 80.34 & 80.32 \\

\bottomrule
\end{tabular}
\end{table}

\subsection{Transfer Learning Analysis}
To evaluate the generalization capabilities of R-DEIM Net, a transfer learning experiment was conducted on seven diverse datasets (Appendix \ref{sec:appendix_data}) spanning community QA, paraphrase identification, and scientific entailment. First, the zero-shot performance of the QQP-trained model directly on these unseen datasets was evaluated, followed by task-specific fine-tuning.

Table \ref{tab:transfer_learning} summarizes these results. In the zero-shot setting, R-DEIM Net demonstrated strong generalization on datasets semantically similar to QQP, such as \textsc{SprintFAQ} (98.15\% $F_1$) and \textsc{SoDD} (63.01\% $F_1$). This indicated that the interaction and reasoning patterns learned from general-domain questions transfer effectively to specific domains without weight updates. 

However, we acknowledge a performance gap on adversarial datasets like \textsc{PAWS} and \textsc{MRPC} in the zero-shot setting. These datasets are specifically designed with high lexical overlap but distinct semantics. We hypothesize that this performance degradation may stem from the Interaction Expert's reliance on localized spatial alignments (e.g., diagonal interaction), which might be bypassed by such adversarial word-scrambling. While further empirical analysis is required to fully confirm this sensitivity to linguistic perturbation, it highlights a potential limitation of efficient spatial models compared to massive cross-attention LLMs.

Fine-tuning further unlocked the potential of the model. Across all datasets, significant performance gains with minimal training (2-4 epochs) were observed. For instance, performance on the challenging \textsc{PAWS} dataset, known for high lexical overlap but distinct semantics, improved from 53.54\% to 67.97\% $F_1$. Similarly, \textsc{CQADupStack} saw a jump from 39.24\% to 80.32\% $F_1$. These results confirm that while R-DEIM Net learns a robust general representation from QQP and remains highly adaptable to specialized nuances.

\subsection{Architectural Analysis}
\label{sec:analysis}

To validate architectural choices, a multi-level analysis was conducted using 10,000 question pairs.

\paragraph{Structural Interaction}
It was hypothesized that duplicate sentence pairs show stronger semantic interaction than non-duplicate sentence pairs, which is reflected by a higher concentration of mass along the leading diagonal of the interaction matrix. Let $M$ be the interaction matrix, $M \in \mathbb{R}^{n \times m}$, where $M_{ij}$ denotes the interaction strength between the $i$-th token of the first sentence and the $j$-th token of the second. Structural interaction was quantified using Eq. \eqref{eq:diag_ratio}.

\begin{equation}
r_{\text{diag}} =
\frac{
\sum_{k=1}^{\min(n,m)} M_{kk}
}{
\sum_{i=1}^{n} \sum_{j=1}^{m} M_{ij}
}
\label{eq:diag_ratio}
\end{equation}

\begin{figure}[t]
    \centering
    \includegraphics[width=0.55\columnwidth]{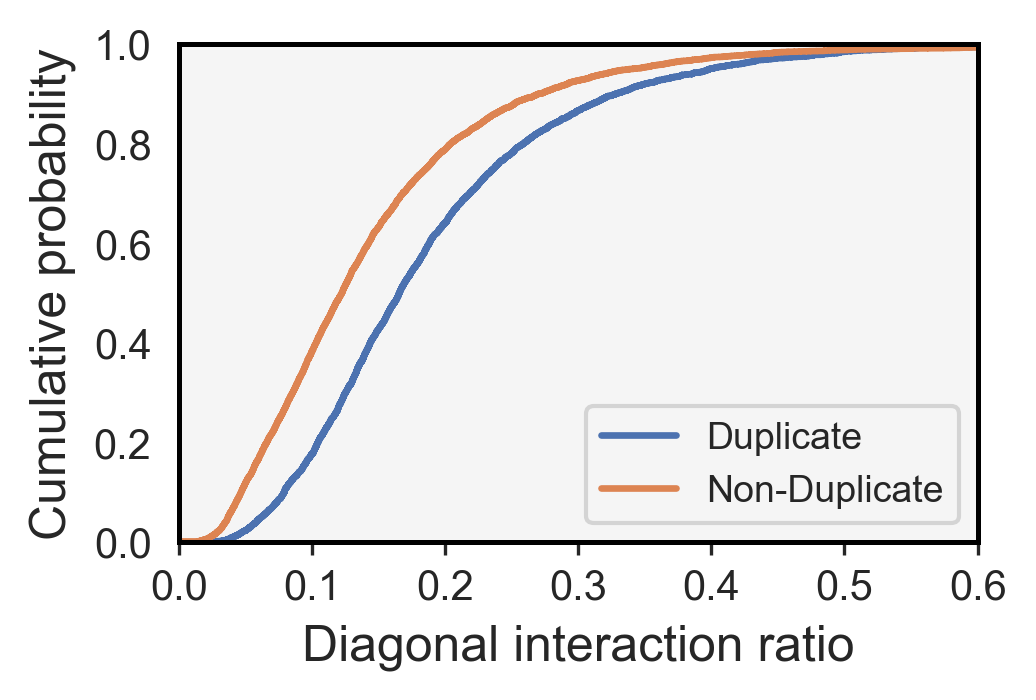}
    \caption{\textbf{Structural Interaction Distribution.} ECDFs of diagonal interaction ratios. The significant rightward shift for duplicates validates that semantic identity is strongly correlated with positional interaction mass.}
    \label{fig:ecdf}
\end{figure}

 The $r_{\text{diag}}$ for all sentence pairs was computed and compared with empirical cumulative distribution functions (ECDFs) between duplicate and non-duplicate examples, as shown in Figure ~\ref{fig:ecdf}. Duplicate sentence pairs exhibited a rightward shift in the ECDF, indicating higher diagonal interaction ratios. To quantify this difference, a two-sample Kolmogorov-Smirnov (KS) test \citep{ks_test} was applied without assuming a specific parametric form. The test revealed a statistically significant distributional difference (KS = 0.224, $p < 10^{-4}$).

Moreover, to assess whether this interaction provides a useful signal, Spearman’s rank correlation was calculated between $r_{\text{diag}}$ and the predicted duplicate probability. A weak positive correlation ($\rho = 0.335$) was observed, indicating that higher interaction is generally associated with higher predicted probability. However, the modest strength of this correlation suggests that interaction alone is insufficient for reliable semantic matching. This observation supports the inclusion of complementary architectural components, such as convolutional and reasoning branches, to handle semantically equivalent but non-aligned sentence pairs.

\paragraph{Multi-Scale Interaction Sensitivity}
Activation magnitudes of the CNN kernels ($3\times3$, $5\times5$, $7\times7$) were examined in Branch A to understand receptive field usage.

As shown in Table \ref{tab:kernels}, the model exhibited a monotonic trend: when diagonal interaction was weak (Low Regime), larger kernels ($5\times5, 7\times7$) exhibited higher relative activation compared to the base $3\times3$ kernel. This indicated that the interaction expert adaptively leverages broader spatial contexts to find semantic matches when they are not diagonally aligned.

\begin{table}[t]
\centering
\caption{Relative activation of larger CNN kernels normalized by the $3\times3$ kernel.}
\label{tab:kernels}

\begin{tabular}{lcc}
\toprule
\textbf{Interaction Regime}
& \textbf{k5/k3}
& \textbf{k7/k3} \\
\midrule
Low  & 1.14 & 1.15 \\
Mid  & 1.11 & 1.09 \\
High & 1.09 & 1.04 \\
\bottomrule
\end{tabular}
\end{table}

\paragraph{Perturbation and Robustness}
To identify linguistic components driving decisions, systematic masking was performed at both token and group levels for 1,000 question pairs.

\begin{figure}[h]
    \centering
    \includegraphics[width=0.8\columnwidth]{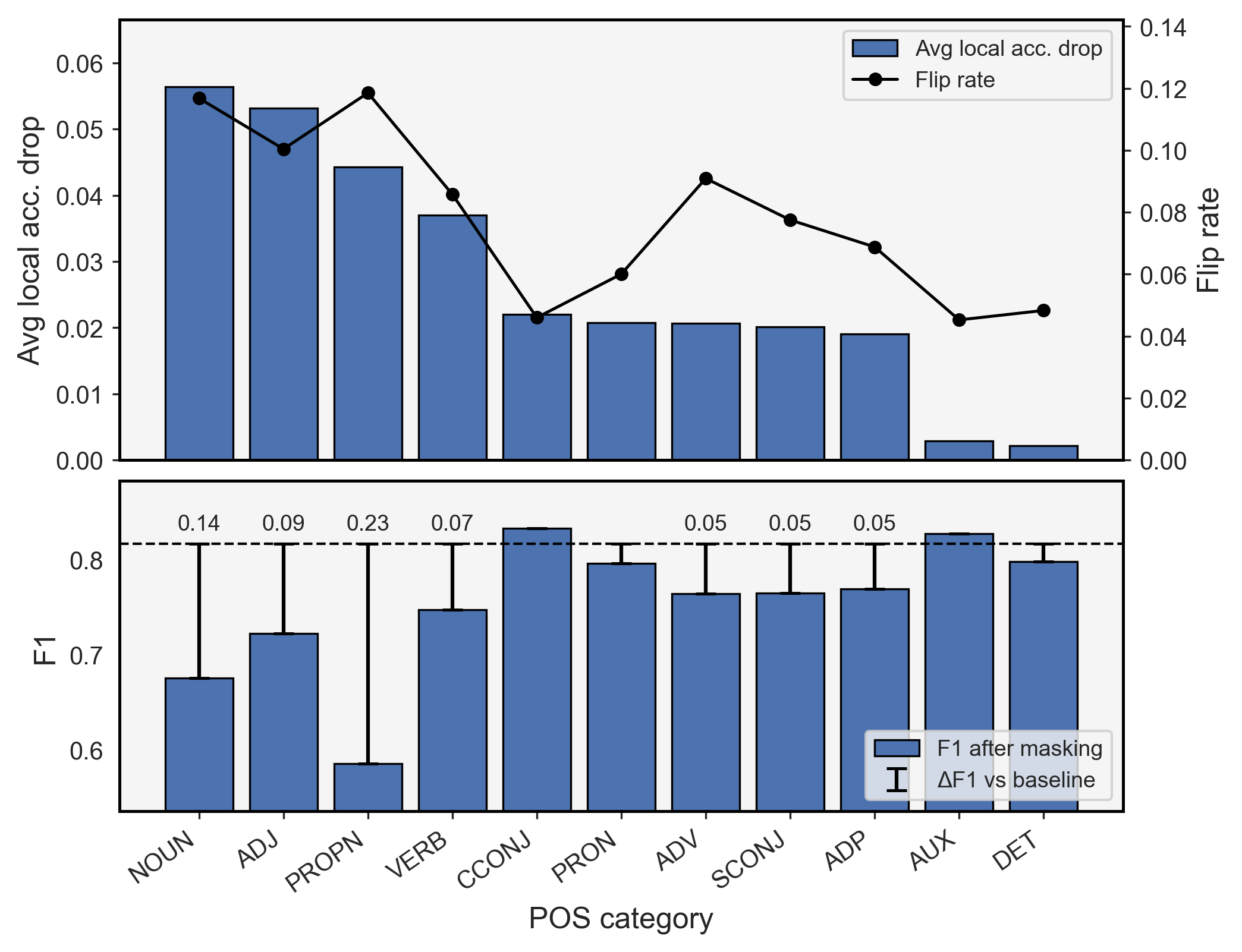}
    \caption{Token-level perturbation analysis. Content words (Nouns, Verbs) exert the strongest influence on model decisions.}
    \label{fig:token_level}
\end{figure}

\subparagraph{Token-Level Sensitivity:} Impact of masking individual tokens is shown in Figure \ref{fig:token_level}. Content-bearing parts of speech, specifically \textsc{Noun}, \textsc{Verb}, and \textsc{Adjective}, induced the highest drops in local accuracy and the highest ``flip rates.'' In contrast, function words like \textsc{Determiners} and \textsc{Auxiliaries} showed minimal impact.

\begin{figure}[t]
    \centering
    \includegraphics[width=0.8\columnwidth]{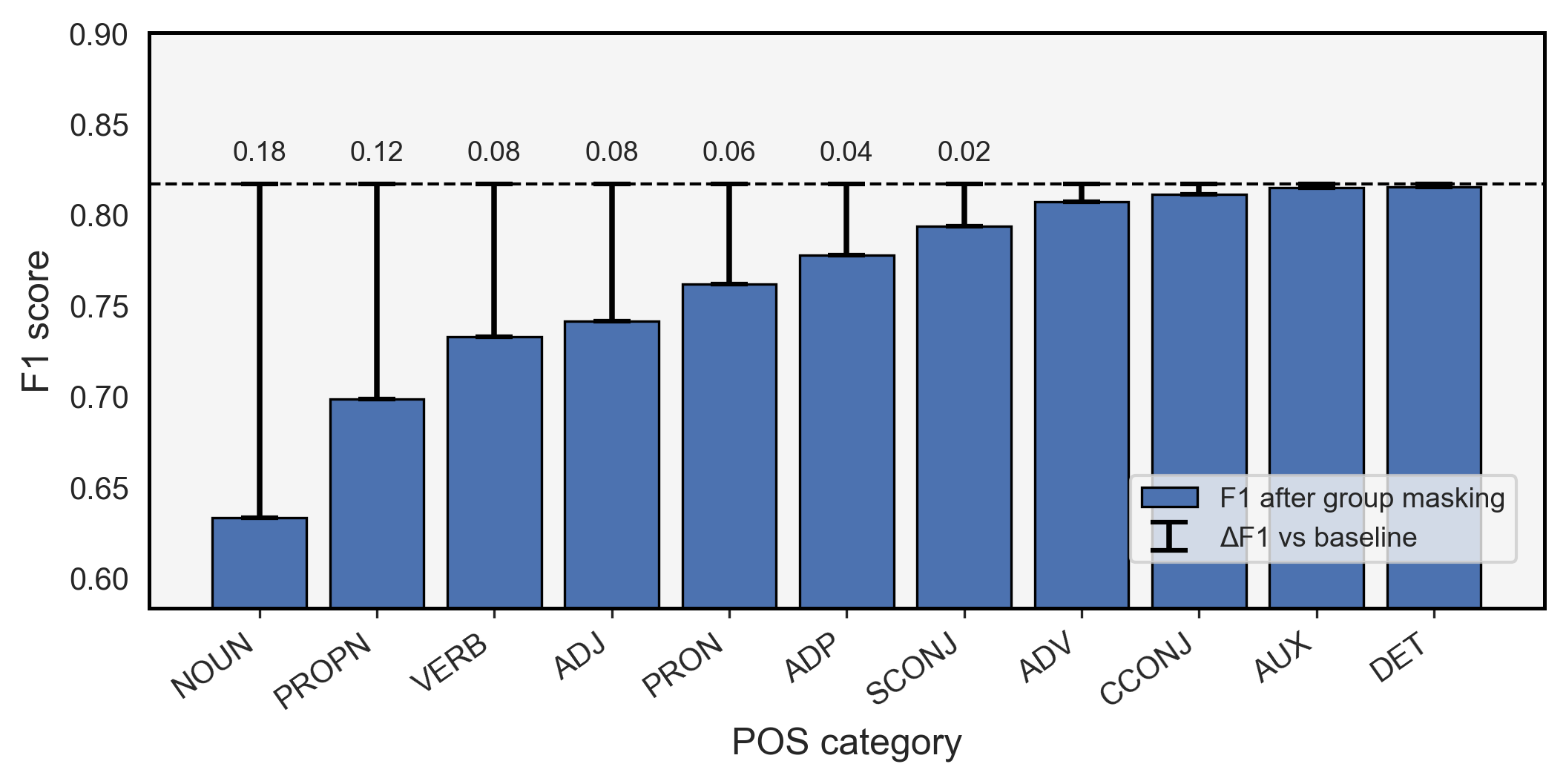}
    \caption{Group-level POS perturbation. The model is highly robust to the removal of syntactic function words but sensitive to the loss of semantic content.}
    \label{fig:group_level}
\end{figure}

\subparagraph{Global Robustness:} Figure \ref{fig:group_level} confirms this trend at the global level. Masking categories of content words led to significant degradation in $F_1$-score, whereas removing function words caused only marginal performance loss. This alignment between local sensitivity and global degradation provides evidence that R-DEIM Net's predictions are grounded in deep semantic content rather than superficial syntactic artifacts.

\section{Conclusion}
\label{sec:conclusion}

R-DEIM Net, a 76M parameter architecture, is presented to resolve the trade-off between massive ``black box'' models and efficient but less accurate alternatives. While recent work achieved high accuracy but poor $F_1$, R-DEIM Net achieved the best of both: 90.07\% accuracy and 90.16\% $F_1$-Score. Ablation results show that incorporating generative reasoning features (Appendix \ref{qualitative_analysis}) yields a consistent improvement over a strong interaction-only baseline, indicating that latent decoder dynamics capture semantic distinctions beyond token interaction alone.
We further validated architectural choices through a multi-level architectural analysis, including structural interaction, CNN kernel activation patterns, and systematic token and POS-level perturbations. These analysis demonstrate that model decisions are driven by semantic content rather than superficial lexical overlap, providing empirical support for the proposed design.

\paragraph{Future Work:}
We will focus on three key areas. First, applying the R-DEIM architecture to other sentence-pair tasks like Natural Language Inference (NLI). Second, investigating transfer learning by testing zero-shot performance on other paraphrase datasets and exploring parameter-efficient fine-tuning. Finally, further miniaturizing the model by exploring distilled decoders to optimize the efficiency-to-performance ratio.

\section*{Limitations}
\label{sec:limitations}
A primary limitation is the use of synthetically generated rationales for the QQP dataset via a large language model. While necessary due to the infeasibility of manually annotating $>$ 400k pairs, this is not ideal compared to human-written ground truth (although our validation study in Appendix \ref{sec:semantic_consistency} indicates strong alignment with human reasoning). However, the strong performance suggests the reasoning vector remains a robust feature, opening avenues for research into whether smaller, human-annotated datasets could yield better results.

Furthermore, while the reasoning vector functionally contributes to classification accuracy, we do not claim strict faithfulness. The generated surface text mimics the structure of LLM-generated rationales (via our training data) but does not definitively prove the model's internal cognitive logic. The text serves as an auxiliary descriptive rationale rather than a guaranteed causal post-hoc explanation. Stronger validation of faithfulness remains an important direction for future work.

In alignment with established SOTA benchmarks on the QQP dataset \citep{70b_sota, bimpm}, the evaluation focuses on \textit{Pair-Level Generalization}. While the QQP dataset contains recurring questions across unique pairs, this experimental design reflects real-world retrieval-based settings where a model must discern semantic identity within unique pair combinations. This approach ensures a direct and fair comparison with existing literature while maintaining high-precision requirements essential for practical duplicate detection.

In this study, computational efficiency and accessibility were prioritized. Experiments were conducted using base-models. While larger models (e.g., RoBERTa-large) might yield incremental performance gains, they require significantly higher VRAM and training time, which were outside the scope of the current hardware infrastructure.

\bibliography{reference_tmlr}
\bibliographystyle{tmlr}
\newpage
\appendix
\section*{Appendix}
\section{Rationale Generation Details}
\label{sec:appendix_rationale}

To facilitate the training of the Reasoning Expert (Branch B), the Quora Question Pairs (QQP) dataset was augmented with synthetic rationales. These rationales were generated using the \texttt{gemini-2.5-flash-lite} model. 

The following system prompt was employed to ensure the generated reasoning was structured and consistent across the dataset, focusing on core intent, entities, and scope while avoiding biased terminology.

\begin{lstlisting}[
    breaklines=true,
    frame=single,
    caption={Prompt used for Synthetic Rationale Generation},
    label={lst:prompt}
]
**Role**: You are a linguistic analyst who provides structured JSON output.

**Task**:
For each question pair provided, analyze it and return a single, valid JSON object.
Your analysis must follow these steps:
1. Identify the Core Intent of each question.
2. Identify the Key Entities in each question.
3. Analyze the Scope of each question.
4. Synthesize these findings into a Final Comment of 30-40 words.
5. Do NOT use the words "duplicate," "same," "different," or "similar" in the final_comment field.

**JSON Output Format**:
{
  "id": <The original ID of the question pair>,
  "core_intent": "Your analysis of the core intent.",
  "key_entities": "Your analysis of the key entities.",
  "scope": "Your analysis of the scope.",
  "final_comment": "Your final synthesized comment."
}

**Example**:
Input:
ID: 123
Question A: "How do I get to the airport?"
Question B: "What's the fastest route to the airport?"

Your Output:
{
  "id": 123,
  "core_intent": "Both questions seek directions to the airport.",
  "key_entities": "The key entity for both is the 'airport'.",
  "scope": "Question B adds a constraint of 'fastest', making it a more specific query than the general request in Question A.",
  "final_comment": "Both inquiries are about finding directions to the airport. One question is a general request for a path, while the other specifically asks for the most time-efficient route available."
}

**Your Turn:**
\end{lstlisting}

\section{Dataset Composition}
\label{sec:appendix_data}

The datasets used in this study vary significantly in scale and label balance. As shown in Table~\ref{tab:dataset_summary}, we evaluated R-DEIM Net on both balanced and highly skewed distributions. Class 0 denotes Non-Duplicates and Class 1 denotes Duplicates.

\begin{table}[h]
\centering
\caption{Dataset statistics and class ratios ($0{:}1$).}
\label{tab:dataset_summary}
\begin{tabular}{lrrc}
\toprule
\textbf{Dataset} &
\textbf{Train ($N$)} &
\textbf{Test ($N$)} &
\textbf{Ratio ($0{:}1$)} \\
\midrule
SODD          & 804,576 & 94,514 & $14{:}5$ \\
SprintFAQ     & 80,800  & 20,200 & $100{:}1$ \\
PAWS          & 49,175  & 2,000  & $63{:}50$ \\
CQADupStack   & 37,924  & 9,482  & $1{:}1$ \\
SciTail       & 23,088  & 2,126  & $17{:}10$ \\
PIT (Twitter) & 11,530  & 838    & $15{:}8$ \\
MRPC          & 3,917   & 1,630  & $12{:}25$ \\
\bottomrule
\end{tabular}
\end{table}

\section{Hyperparameters}
\label{sec:Hyperparamters}
Table \ref{tab:hyperparameters} outlines the exact hyperparameter configurations used to train the components of R-DEIM Net, ensuring reproducibility across all experimental setups documented in Section 4.

\begin{table}[h]
\centering
\caption{Hyperparameters and configuration of R-DEIM Net.}
\label{tab:hyperparameters}

\begin{tabular}{ll}
\toprule
\textbf{Parameter} & \textbf{Value} \\
\midrule
Shared Encoder & \texttt{MiniLM-L12-v2} (384-dim) \\
SLM Decoder & \texttt{Flan-T5-small} (512-dim) \\
CNN Out Channels (per kernel) & 128 \\
CNN Kernel Sizes & [3, 5, 7] \\
ANN Hidden Size & 512 \\
ANN Dropout & 0.3 \\
Batch Size & 32 \\
Epochs & 5 \\
Max Length (T5) & 32 tokens \\
BERT Learning Rate & $2\times10^{-5}$ \\
SLM Learning Rate & $5\times10^{-5}$ \\
Custom Heads Learning Rate & $1\times10^{-4}$ \\
Multi-Task Loss Alpha ($\alpha$) & 0.7 \\
BCE False Positive Penalty & 2.0 \\
Importance Factor ($b_{imp}$) & 0.01 \\
Cross-Validation & 10-fold \\
Random State (Global) & 42 \\
\bottomrule
\end{tabular}
\end{table}

\section{Semantic Consistency and Human Evaluation}\label{sec:semantic_consistency}To validate the semantic consistency of the synthetically generated rationales, an embedding-based alignment experiment was conducted. Question pairs were concatenated using a separator token (i.e., \texttt{Q1 + [SEP] + Q2}) and processed through a pre-trained encoder (\texttt{paraphrase-MiniLM-L12-v2}) to obtain unified pair embeddings. An identical encoding operation was performed on the LLM-generated ``Final Comment'' rationales. We then computed the cosine similarity between the question pair embeddings and their corresponding rationale embeddings. For correctly matched pairs, the mean cosine similarity was 0.5670. To establish a baseline, rationales were also paired with randomly selected question pairs, which caused the mean similarity to drop sharply to 0.0662. This distinct margin confirms that the generated rationales are highly specific and semantically grounded in their respective input pairs and not relying on generic formulations.

Additionally, a human evaluation study was conducted with 12 annotators on a subset of over 170 randomly selected pairs to verify the quality of the generated text. Annotators were tasked with providing ``ground-truth'' reasoning for the semantic relationship between the questions. These human-written rationales were then compared against the model-generated rationales using embedding similarity. The results demonstrated strong alignment, yielding a Spearman Correlation of 0.814 and a Cosine Similarity of 0.763. The descriptive depth was also comparable, with an average token length of 24.50 for human-written rationales and 28.02 for the Gemini-generated rationales. To further illustrate this alignment, the top examples identified by the correlation metric ($\rho \approx 0.88 - 0.89$) highlight how closely the synthetic rationales mimic human analysis:

\begin{itemize}
    \item \textbf{Example A (Correlation: 0.891):}
    \begin{itemize}
         \item \textit{Human-Written:} ``The first question is asking about what defines secondary education in India while the second question is asking about whether Archie Comics should advertise more.''
         \item \textit{Generated:} ``These questions address disparate topics. One seeks a definition of secondary education in India, while the other inquires about promotional efforts for Archie Comics.''
    \end{itemize}
     \item \textbf{Example B (Correlation: 0.883):}
    \begin{itemize}
         \item \textit{Human-Written:} ``Both questions show the intent of the person asking this question about the famous movie series Harry Potter and the reasoning behind it. They are asking for subjective opinions.''
         \item \textit{Generated:} ``Both questions seek a personal favorite from the Harry Potter film series and the underlying reasons for that selection. They are asking for a subjective opinion with supporting rationale.''
    \end{itemize}
\end{itemize}

\section{Qualitative Analysis}
\label{qualitative_analysis}
As shown in Figure \ref{fig:qualitative}, for true duplicates (Example 1), the model successfully articulates the shared intent. Conversely, in Example 2, the Reasoning Expert explicitly contrasts the differing constraints (salary vs. skills), offering a natural language description of the semantic relationship alongside the prediction. For true duplicates, R-DEIM Net confidently predicts the class and articulates the shared intent. For non-duplicates, it correctly identifies the distinction, proving it relies on semantic understanding rather than simple keyword overlap.

\begin{figure}[h]
    \centering
    \fbox{
        \parbox{1\columnwidth}{
        \textbf{Example 1 (True Duplicate):} \\
        \textbf{Q1:} How I can improve my English communication? \\
        \textbf{Q2:} How can I improve English speaking skill?\\
        \textbf{Prediction:} Duplicate (0.9620) \\
        \textbf{Generated rationale:} Both questions are seeking advice on enhancing English language skills. One is a request for improvement in English communication, while the other targets the improvement of spoken English.
 \\
        \textbf{Example 2 (Non-Duplicate):} \\
        \textbf{Q1:} What is the average salary of a data scientist in London? \\
        \textbf{Q2:} What skills do I need to become a data scientist? \\
        \textbf{Prediction:}  non-duplicate (0.0965) \\
        \textbf{Generated rationale:} Both questions are seeking information regarding data scientist domain. One is asking about the salary of data scientist in London, while the other is asking about the skills.}
        
    }
    \caption{\textbf{Qualitative Analysis.} Sample outputs showing calibrated duplicate probabilities alongside human-readable rationales that justify semantic identity or distinction.}
    \label{fig:qualitative}
\end{figure}

\end{document}